# Interpretable Disease Prediction based on Reinforcement Path Reasoning over Knowledge Graphs

Zhoujian Sun, Wei Dong, Jinlong Shi and Zhengxing Huang

*Abstract*— **Objective: To combine medical knowledge and medical data to interpretably predict the risk of disease. Methods: We formulated the disease prediction task as a random walk along a knowledge graph (KG). Specifically, we build a KG to record relationships between diseases and risk factors according to validated medical knowledge. Then, a mathematical object walks along the KG. It starts walking at a patient entity, which connects the KG based on the patient's current diseases or risk factors and stops at a disease entity, which represents the predicted disease. The trajectory generated by the object represents an interpretable disease progression path of the given patient. The dynamics of the object are controlled by a policy-based reinforcement learning (RL) module, which is trained by electronic health records (EHRs). Experiments: We utilized two real-world EHR datasets to evaluate the performance of our model. In the disease prediction task, our model achieves 0.743 and 0.639 in terms of macro area under the curve (AUC) in predicting 53 circulation system diseases in the two datasets, respectively. This performance is comparable to the commonly used machine learning (ML) models in medical research. In qualitative analysis, our clinical collaborator reviewed the disease progression paths generated by our model and advocated their interpretability and reliability. Conclusion: Experimental results validate the proposed model in interpretably evaluating and optimizing disease prediction. Significance: Our work contributes to leveraging the potential of medical knowledge and medical data jointly for interpretable prediction tasks.**

## I. Introduction

Disease prediction is one of the most important research topics in medical informatics because of its potential for forecasting the prognosis of patients, reducing preventable adverse events, and decreasing medical costs [1]-[3]. Recently, benefiting from the development of machine learning (ML) technology and accessibility of electronic health records (EHR), modern ML-based disease prediction models have achieved impressive progress in prediction performance compared to traditional statistical models [4]. However, the application perspective of ML-based models is still unclear because most of them are black boxes, and therefore their decision-making procedures are difficult to understand by clinicians. This limitation raises concerns about the reliability of the ML-based models and inevitably restricts their deployment in clinical practice [5].

To address this problem, researchers have proposed many interpretable disease prediction models [6]. These models typically are able to generate a set of weights to indicate the contribution of each feature to the prediction as explanations. For example, Choi *et al.* proposed an attention mechanism-based model to predict the onset of heart failure [7], Das *et al.* proposed a fuzzy rule-based model to predict the onset of Alzheimer's disease [8], and Bernardini *et al.* used a revised support vector machine (SVM) to predict the onset of diabetes [9]. Although valuable, we find that these models still cannot obtain sufficient trust from clinicians because they interpret predictions in a purely data-driven manner without utilizing any prior medical knowledge. Note that, because modern evidence-based medicine (EBM) demands that all clinical decisions should be made under clinical evidence [10], clinicians typically expect machine learning models to leverage medical knowledge and medical data comprehensively to conduct reliable, interpretable predictions [5]. To date, this challenge has not been well addressed. Although several studies have attempted to utilize medical knowledge to improve disease prediction performance, the models proposed in these studies are still unexplainable to some extend [11]-[14].

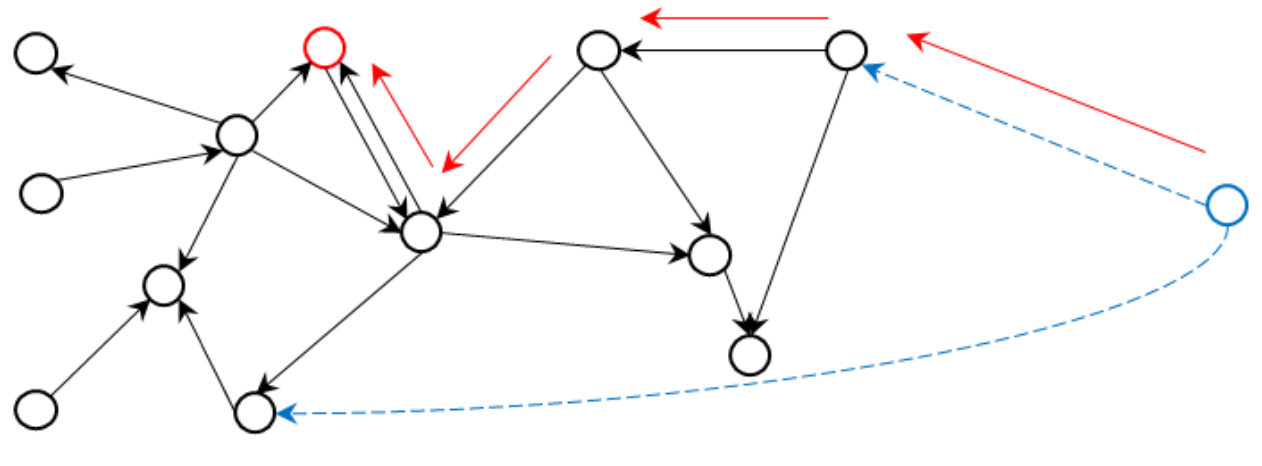

Predicted Disease Entity
Patient Entity
KG Entity
Disease Progression Path
Patient-KG connection

Fig. 1 Interpretable disease prediction over a knowledge graph. A patient entity connects the KG according to its diseases or risk factors. Then, a mathematical object controlled by an RL agent starts walking at the patient entity and stop walking at a disease entity. The terminated disease entity is the predicted disease, and the object's trajectory indicates the interpretable disease progression path.

In this study, we have adopted a new perspective that reformulates the interpretable disease prediction task into a random walk along a knowledge graph (KG) to address this challenge (Figure 1). Under this perspective, we propose a model that consists of a KG and a reinforcement learning (RL) module. The KG records the validated medical knowledge, where an arbitrary path indicates a possible disease progression path. We then establish connections between the KG and a patient entity based on its current diseases or risk factors. After that, a hypothetical mathematical object starts walking from the patient entity and stops at an entity after several steps. The terminated entity is the predicted disease, and the trajectory is the interpretable disease progression path of the predicted disease. To ensure that the mathematical object stops at a target entity, i.e., the disease that indeed occurs in the future, as much as possible, we train a policy-based RL module to control the dynamics of the object. Specifically, we construct a state vector at each timestep by summarizing the patient feature, current entity, and historical trajectory. The RL module can generate a policy vector according to the state vector to guide the transition of the object. Finally, if the object stops walking at the target entity, the RL module obtains rewards and updates its parameters. Therefore, we are able to optimize the RL module by utilizing patient medical data. Once the RL module has been optimized, our model can predict diseases with high occurrence probabilities and generate interpretable disease progression paths simultaneously. We applied two real-world EHR datasets to evaluate the performance of the proposed model. The experimental results demonstrated that our model is comparable to commonly seen machine learning approaches utilized in clinical studies. Meanwhile, the reliability of disease progression paths generated by our model is reviewed and advocated by our clinical collaborator.

Our main contributions are as follows.

1) We present a novel KG-based learning model for the prediction of disease. To the best of our knowledge, this is the first study that a disease prediction problem has been interpretably addressed by incorporating validated medical knowledge into a data-driven model.

2) We propose equipping a neural network with a KG to learn representations from a large volume of EHR data. The results prove that the generated knowledge path considerably improves the interpretability of the disease prediction.

3) Our approach is extensively evaluated in two real-world clinical datasets. The experimental results demonstrate that our model is as effective as state-of-the-art machine learning methods that are widely used in clinical research. Professional clinicians also advocated the reliability of interpretable disease progression paths generated by the model.

The source code and the scripts to build the KG in the Neo4j database are available at https://github.com/ZJU-BMI/PBXAI.

## II. Related Work

### A. Interpretable Disease Prediction Model

From a technical perspective, existing interpretable disease prediction models can be coarsely separated into four subtypes [6]. 1) Attention-based methods. The attention-based models leverage a neural network to assign each feature a sample-specific weight as the explanation. For example, Choi *et al.* [7] and Yin *et al.* [15] adopted attention mechanisms and the recurrent neural network (RNN) to predict the onset of heart failure in an explainable manner, respectively. 2) Model-agnostic methods, with Local Interpretable Model-agnostic Explanations (LIME) [16] and SHapley Additive exPlanations (SHAP) [17] being two representative approaches. A model-agnostic method is capable of endowing interpretability to any unexplainable model by reorganizing the prediction result into a linear combination of features of the given sample. For example, Tseng *et al.* utilized a random forest to predict acute kidney injury, and they adopted SHAP to interpret the prediction as a linear combination of given features [18]. Yan *et al.* used a convolutional neural network (CNN) to predict age-related macular degeneration, and they adopted LIME to interpret the prediction [19]. 3) Knowledge distilling method. It utilizes an accurate yet unexplainable method as a supervisor to train a light, explainable model. The trained light model can achieve similar performance compared to the supervisor without the loss of interpretability. For example, Das *et al.* adopted a weighted sum of fuzzy rules to explain the prediction of Alzheimer's disease that was conducted by an unexplainable kernel-based method [8], and Che *et al.* leveraged the gradient boosting tree to explain the prediction of mortality that was conducted by an RNN [20]. 4) Revising intrinsically interpretable models. For example, Brisimi *et al.* [21] and Bernardini *et al.* [9] revised SVM to predict patient readmissions and onset of diabetes, respectively. In summary, the interpretability of existing ML-based disease prediction models relies on assigning each feature a specific importance weight (or score) to indicate its contribution to the prediction. All of these four types of methods are purely data-driven and fail to utilize existing medical knowledge to provide interpretable predictions.

### B. Prediction Model Leveraging Medical Knowledge

As far as we know, existing ML studies mainly leverage medical knowledge to learn efficient representations of medical concepts. The learned medical concept representations can improve the performance of downstream tasks, e.g., disease risk prediction. For example, Choi *et al.* proposed GRAM, which utilizes the hierarchical structure of the international classification of disease (ICD) encoding system and attention mechanism to learn the representations of disease concepts [11], and then applied the learned representations in the disease prediction task. The experimental results of three independent studies demonstrated that this method could significantly improve the performance in predicting the onset of rare diseases [11], [12], [14]. Zhang *et al.* and Yin *et al.* utilized the TransE and graph attention network to learn representations of medical concepts from a KG named KnowLife [22], and they applied the learned representations in predicting the onset of heart failure [13], [23]. Although these studies demonstrated the efficacy of introducing medical knowledge, they did not propose transparent models whose decision-making process is understandable to clinicians, which we will solve in this study.

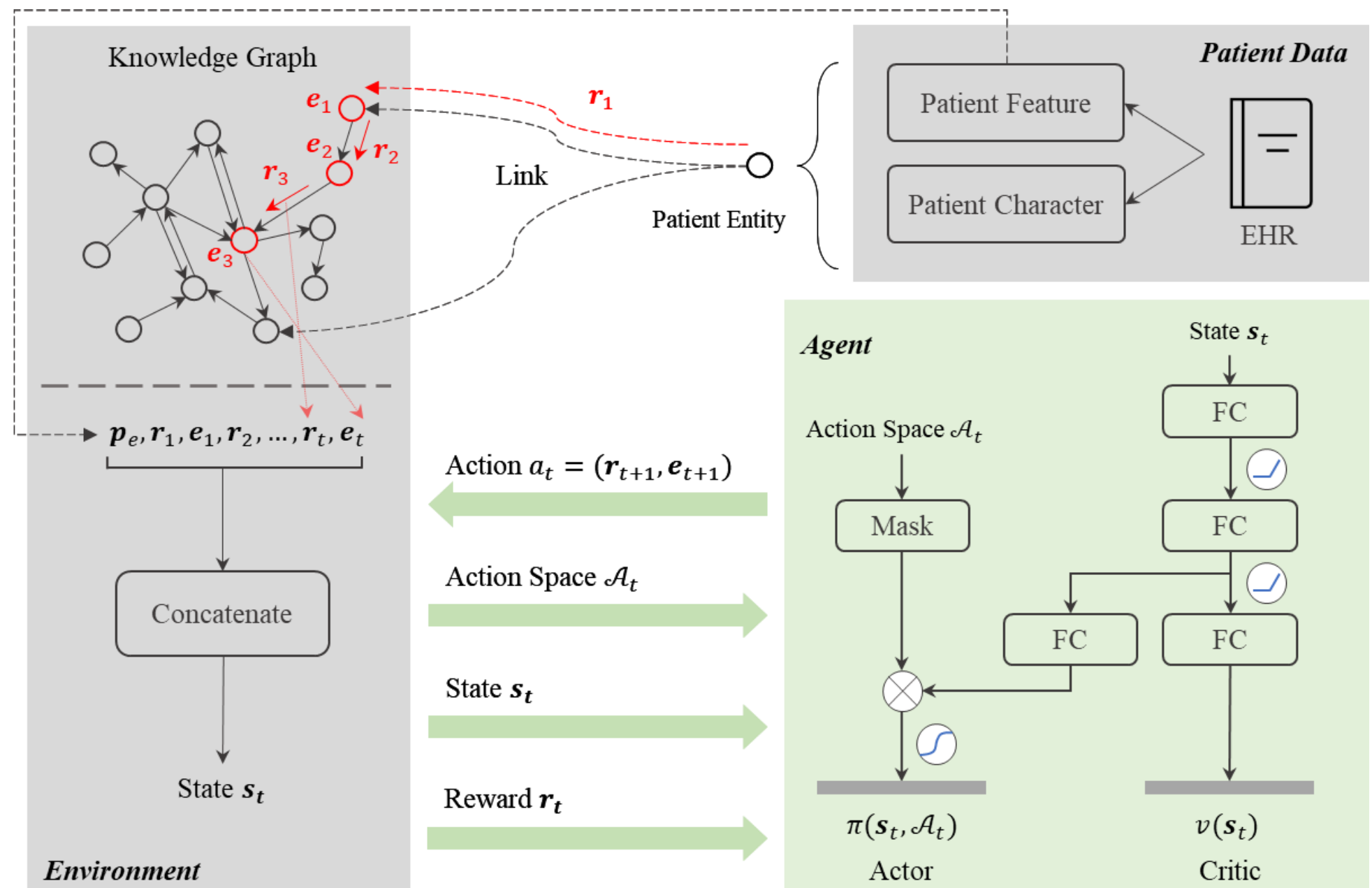

Fig. 2 Model Framework. We connect the patient entity and the KG according to the patient characters $p_c$. An actor-critic RL module is used to control a hypothetical mathematical object to traverse the KG. We feed the state $\boldsymbol{s}_t$, the concatenation of the patient feature, current entity, and history trajectory, to the RL module, which generates both a policy vector $\pi$ and the estimated state value $v$.

### C. *Path-Based Interpretable Recommender*

The path-based method is an emerging interpretable method in recommendation systems [24]. For example, Xian *et al*. [25] and Zhang *et al*. [26] have utilized this method to achieve interpretable, individualized recommendations for customers. The core of the path-based method is building a large graph, referred to as a heterogeneous information network (HIN), to record all customers, products, reviews, and their relations. After that, products are recommended for customers by generating recommend paths along the HIN. Existing path-based method studies generally utilize an RL agent that is trained by a large volume of purchase data to generate the recommended products, as well as the recommendation paths [27], [28]. Because the paths have explicit semantic explanations, the path-based method is regarded as an interpretable method.

Inspired by the path-based method, we built a KG to record validated medical knowledge and connect patients to the KG according to their diseases and risk factors in this study. After that, we use EHR data to train an RL agent. The trained RL agent can generate predicted diseases and disease progression paths along the KG as explanations. It is worth noting that we are not the first to introduce the path-based method in medical research. Xu *et al*. first leveraged the path-based method in disease prediction [29]. However, their model did not leverage validated medical knowledge and cannot predict diseases for unseen patients either. In this study, we carefully addressed these defects to ensure that our model has the potential to be deployed in a real clinical environment.

## III. METHODOLOGY

### A. *Preliminaries*

We define a KG $\mathcal{G}$ with an entity set $\mathcal{E}$ and a relation set $\mathcal{R}$ as $\mathcal{G} = \{(e_h, r, e_t) | e_h, e_t \in \mathcal{E}, r \in \mathcal{R}\}$, $|\mathcal{E}| = m, |\mathcal{R}| = n$ . Each entity in $\mathcal{E}$ represents a medical concept, e.g., coronary heart disease, diabetes, overweight, etc. Each triplet $(e_h, r, e_t)$ indicates the knowledge of the relation $r$ from the head entity $e_h$ to the tail entity $e_t$, e.g., "myocardial infarction" (head entity) "causes" (relation) "heart failure" (tail entity).

To create connections between a patient entity $p$ and the KG $\mathcal{G}$, we reorganize the raw EHR data of $p$ into two parts, patient characters $\boldsymbol{p}_c \in \{0,1\}^m$ and patient features $\boldsymbol{p}_f \in \mathbb{R}^l$ (Figure 2). Each element in $\boldsymbol{p}_c$ can be directly mapped to a corresponding entity in $\mathcal{G}$. For example, if a patient $p$ has hypertension and heart failure, the corresponding elements in $\boldsymbol{p}_c$ will be set to 1. Thereafter, we can establish connections between the patient entity and KG according to the element value of $\boldsymbol{p}_c$. We define the relations that connect the patient entity, and the KG entity has the relation type "have". Patient features $\boldsymbol{p}_f$ include information that is useful to the prediction task but cannot be mapped to the KG in the EHR dataset, e.g., lab test results, demographics, vital signs, etc.

We formulate the prediction task as a random walk process. Specifically, we let a mathematical object start walking at the patient entity and stop walking after T steps. The goal of our study is to make the object stop at a target disease entity that patient $p$ indeed develops in the future as much as possible. Note that the mathematical object may reach the target entity at step t (t<T). To allow the object to stuck at the target entity rather than continue past the target entity, we add self-loops to all entities in the KG. In the following subsections, we denote the graph that consists of KG $\mathcal{G}$, self-loops, patient entity $p$, and related patient-KG connections as $\mathcal{G}_c$.

## *B. Reinforcement Learning Module*

We utilized the actor-critic RL-based framework [30] to control the dynamics of the mathematical object and generate the disease progression path.

### *1) State*

The state $\boldsymbol{s}_t$ at timestep $t$ is defined as $\boldsymbol{s}_t = (\boldsymbol{p}_e, \boldsymbol{e}_t, h_t)$, where $\boldsymbol{e}_t$ are representations of an entity at timestep $t$ and $\boldsymbol{p}_e$ is the patient representation learned from $\boldsymbol{p}_f$. The details of generating representations of patients and entities will be described in the next subsection. $h_t$ is the historical trajectory before step $t$. In this study, we use the 1-step history to represent the influence of history, i.e., $h_t = (\boldsymbol{e}_{t-1}, \boldsymbol{r}_t)$. Conditioned on $\boldsymbol{p}_e$, the initial state is defined as $\boldsymbol{s}_0 = (\boldsymbol{p}_e, \boldsymbol{p}_e, \boldsymbol{0})$. Given the fixed horizon $T$, the terminal state $\boldsymbol{s}_T$ is defined as $s_0 = (\boldsymbol{p}_e, \boldsymbol{e}_T, h_T)$.

### *2) Actor-Critic Module*

We utilize a two-layer fully connected (FC) neural network to learn a representation $\boldsymbol{x}_t$ from $\boldsymbol{s}_t$ and then design a policy network for learning policy $\boldsymbol{\pi}_t$ and a value network for learning value $v_t$ according to $\boldsymbol{x}_t$. Equations 1-3 describe the process of generating $\boldsymbol{\pi}_t$ and $v_t$.

$$\boldsymbol{x}_t = \sigma(\sigma(\boldsymbol{s}_t \boldsymbol{W}_1 + \boldsymbol{b}_1)\boldsymbol{W}_2 + \boldsymbol{b}_2) \quad (1)$$

$$\boldsymbol{\pi}_t = \mathrm{softmax}\big((\boldsymbol{x}_t \boldsymbol{W}_p + \boldsymbol{b}_p) \odot \boldsymbol{\mathcal{A}}_t\big) \quad (2)$$

$$v_t = \boldsymbol{x}_t \boldsymbol{W}_v + \boldsymbol{b}_v \quad (3)$$

Here, $\sigma$ is a nonlinear activation function, for which we use the rectified linear unit (ReLU) in this study, $\odot$ indicates the Hadamard product, and $\{\boldsymbol{W}_1, \boldsymbol{W}_2, \boldsymbol{W}_p, \boldsymbol{W}_v, \boldsymbol{b}_1, \boldsymbol{b}_2, \boldsymbol{b}_p, \boldsymbol{b}_v\}$ are the parameters and bias in the neural network.

The action space $\mathcal{A}_t$ describes the set of available actions, which includes all outgoing edges, as well as the related tail entities, of $e_t$. An action $a \in \mathcal{A}_t$ is defined as a tuple of an outgoing edge and the corresponding tail entity. Meanwhile, to avoid the mathematical object reaching an entity repeatedly, we excluded all actions that point to the historical entities in practice (except the self-loop action). Formally, $\mathcal{A}_t$ can be defined as:

$$\mathcal{A}_t = \{(r, e) | (e_t, r, e) \in \mathcal{G}_c, e \notin \{p, e_1, e_2, \ldots, e_{t-1}\}\} \quad (4)$$

Of note, a primary challenge of our task, compared to the commonly seen RL learning task, is that the available action space varies in every state because the outgoing edges set of each entity in $\mathcal{G}_c$ is different. To address this challenge, we reorganize $\mathcal{A}_t$ into an m-dimensional mask vector $\boldsymbol{\mathcal{A}}_t$, where each element indicates whether the corresponding entity is reachable in $\mathcal{A}_t$. If an entity is unreadable at step $t$, its logits will be forced to negative infinity to ensure its probability will be assigned to zero (Equation 2), and we can guarantee that only the available transition can be sampled according to $\pi(s_t, \mathcal{A}_t)$ in the random walk.

### *3) Reward*

To encourage the mathematical object to terminate at a disease entity, we set the reward $R_T$ as follows:

$$R_T = \begin{cases} 1, \text{if } \mathrm{e_T} \in \text{disease and occurs in the future} \\ 0, \text{if } \mathrm{e_T} \text{ is a disease but does not occur in the future} \\ -1, \text{otherwise} \end{cases} \quad (5)$$

**Algorithm** 1: Disease Prediction and Path Reasoning

**Input**: patient representation $\boldsymbol{p}_e$, time horizon $T$, Knowledge Graph $\mathcal{G}_c$, patient entity $p$, beam search set $[K_0, K_2, \ldots, K_{T-1}]$, and disease number $D$

**Output**: disease probability set $P$, path set $\mathcal{R}_T$, and path probability set $\mathcal{Q}_T$

1 Initialize: $\mathcal{R}_0 \leftarrow \{\{p\}\},\ \mathcal{Q}_0 \leftarrow \{1\},\ P \leftarrow \{0\}^D$
2 **for** $t \leftarrow 0$ **to** $T-1$ **do**:
3 Initialize $\mathcal{R}_{t+1} \leftarrow \emptyset, \mathcal{Q}_{t+1} \leftarrow \emptyset$
4 **forall** $\hat{r} \in \mathcal{R}_t, \hat{q} \in \mathcal{Q}_t$ **do**:
5 Path $\hat{r} \doteq \{p, r_1, e_1, \ldots, r_t, e_t\}$;
6 Set $\boldsymbol{s}_t \leftarrow (\boldsymbol{p}_e, \boldsymbol{e}_t, h_t)$ // refer to Section III.B.1
7 // refer to Equation 4
Get action space $\mathcal{A}_t$ according to $\mathcal{G}_c$, $\hat{r}$, and $e_t$
8 // $f(\cdot)$ is the combination of Equation 1,2
Generate policy: $\boldsymbol{\pi}_t = f(\boldsymbol{s}_t, \mathcal{A}_t)$
9 Delete actions whose probabilities are not in top $K_t$
10 **forall** $a \in \mathcal{A}_t$ **do**:
11 Save new path $\hat{r} \cup (r_{t+1}, e_{t+1})$ to $\mathcal{R}_{t+1}$`
12 Save new probability $\pi_t(a) \times \hat{q}$ to $\mathcal{Q}_{t+1}$
13 **end**
14 **end**
15 **end**

16 **forall** $\hat{r} \in \mathcal{R}_T, \hat{q} \in \mathcal{Q}_T$ **do**:
17 Path $\hat{r} \doteq \{p, r_1, e_1, \ldots, r_T, e_T\}$
18 **if** $e_T$ is a disease entity:
19 Get the index $idx$ of element $e_T$
20 $P[idx] \leftarrow P[idx] + \hat{q}$
21 **end**
22 **end**
23 **return** $P, \mathcal{P}_T, \mathcal{Q}_T$

The rewards of all other time steps are set to zero, i.e., $\mathrm{R_t} = 0$ for all $t<T$.

### *4) Regularization and Optimization*

The goal of optimization is to learn a stochastic policy $\pi$ that can achieve the largest expected cumulative expected rewards.

$$J(\theta) = \mathbb{E}_\pi \left[ \sum_{t=0}^{T-1} \gamma^t R_{t+1} \right] \quad (6)$$

Here, $\theta$ represents all parameters in the model, and $\gamma$ is the decay coefficient.

Of note, the data-imbalance problem is common in medical studies. Most patients suffer from several types of diseases, and samples of other diseases are rare. This characteristic may cause our model to sample the path point to the common disease entities repeatedly, e.g., hypertension, diabetes, and angina, and neglect other diseases. To encourage the RL agent to explore diverse paths, we add a regularization term $H(\pi)$ representing the entropy of policy. Finally, according to the optimization method of the actor-critic framework, the model gradient with the entropy term $J'(\theta)$ is defined in Equation 7, from which we can use a gradient-based method to optimize the parameters in the proposed model.

$$\nabla_\theta J'(\theta) = \mathbb{E}_\pi \big[ \nabla_\theta \log \pi(\cdot\,|s, \boldsymbol{\mathcal{A}}) \big( G - v(\boldsymbol{s}) \big) \big] + \alpha \nabla_\theta H(\pi) \quad (7)$$

Here, $G$ is the discounted cumulative reward from the initial step to the terminal state $s_T$, and $\alpha$ is the weight of the entropy term.

### *5) Inference*

Once the parameters in the model are optimized, we can predict diseases and generate progression paths for any given patient. The process is described as Algorithm 1. It takes the RL agent, $\mathcal{G}_c$, time horizon $T$, number of predicted diseases $D$, patient $p$, and $\boldsymbol{p}_e$ as inputs. As the number of possible progression paths explodes with the increase of $T$, we employ beam search to only take actions with top-$K_t$ probabilities at step t into consideration. The block from line 4 to line 14 is the core of the algorithm. We will generate policy $\boldsymbol{\pi}_t$ according to the current state (from line 5 to line 8). Then, the trimmed action space $\mathcal{A}_t$ is generated, where actions with small probabilities are deleted (line 9). Thereafter, traverse all possible actions to generate the path set $\mathcal{R}_t$ and path probability set $Q_t$ at timestep t (line 10 to line 13). After $T$ loops, we can obtain path set $\mathcal{R}_T$ that contains all possible disease progression paths and the related probability set $Q_T$. Finally, the probability of each disease is summarized (line 16 to line 22). Note that there may exist multiple disease progression paths between the patient $p$ and a disease entity $e$. We will reserve all paths to interpret the possible disease progression paths of the given patient.

### *C. Representation Learning of Entities and Relations*

We describe how to learn $\boldsymbol{p}_e$, $\boldsymbol{e}_i$, and $\boldsymbol{r}_i$ in this subsection. The representations of patient, entity, and relation are learned independently before running the RL model.

#### *1) Entity Representation*

Inspired by Tran *et al*. [31], we adopted the restricted Boltzmann machine (RBM) to learn the representation of entities in KG. Specifically, given a patient character vector $\boldsymbol{p}_c$ and parameters $\boldsymbol{W}, \boldsymbol{a}, \boldsymbol{b}$, the data likelihood $p(\boldsymbol{p}_c)$ follows:

$$p(\boldsymbol{p}_c) = \sum_{\boldsymbol{h}} p(\boldsymbol{p}_c, \boldsymbol{h}) \tag{8}$$

$$p(\boldsymbol{p}_c, \boldsymbol{h}) \propto \exp(\boldsymbol{a}^T \boldsymbol{p}_c + \boldsymbol{b}^T \boldsymbol{h} + \boldsymbol{p}_c^T \boldsymbol{W} \boldsymbol{h}) \tag{9}$$

where $\boldsymbol{h} \in \{0,1\}^k$, and the parameters can be optimized by maximizing $p(\boldsymbol{p}_c)$ [32]. Once the parameters are optimized, the elements of the row vector $W_{i\star}$ coordinate the representation of an entity that can be mapped to the i[th] element of $\boldsymbol{p}_c$.

#### *2) Patient Representation*

We used a two-layer autoencoder to learn the patient representation $\boldsymbol{p}_e$. In particular, given a patient feature vector $\boldsymbol{p}_f$, the distributed representation $\boldsymbol{p}_e$ follows $\boldsymbol{p}_e = f(\boldsymbol{W_1} f(\boldsymbol{W_2} \boldsymbol{p}_e))$, where $\boldsymbol{W_1}, \boldsymbol{W_2}$ are the parameters, and $f(\cdot)$ is a nonlinear function. The parameters are learned by minimizing binary cross-entropy.

$$\mathcal{L} = \sum_i \boldsymbol{p}_{f_i} \log \hat{\boldsymbol{p}}_{f_i} + (1 - \boldsymbol{p}_{f_i}) \log(1 - \hat{\boldsymbol{p}}_{f_i}) \tag{10}$$

Here, $\hat{\boldsymbol{p}}_f = f(\boldsymbol{W}_2^T f(\boldsymbol{W}_1^T \boldsymbol{p}_e))$ is the reconstructed $\boldsymbol{p}_f$.

#### *3) Relation Representation*

In this study, we use one-hot embedding to represent different relations.

## IV. Experiments

### *A. Knowledge Graph*

We manually constructed a KG concerning circulation system diseases in this study based on an authoritative medical textbook [33]. The proposed KG includes 65 entities (53 disease entities, five disease category entities, and seven risk factor entities), covering all commonly seen circulation system diseases and related risk factors. We also identified 326 relations to describe relationships between entities comprehensively. The quality of the KG was reviewed and approved by our clinical collaborator, a director of the cardiology department of an academic tertiary medical center. The KG is shown in Supplementary Figure 1.

TABLE 1
Statistics of Preprocessed Data

| Statistics | PLAGH | MIMIC-III |
|---|---|---|
| # of patients | 9927 | 6446 |
| # of hospitalizations | 30918 | 10693 |
| # of features in $\boldsymbol{p}_f$ | 41 | 33 |
| Avg. # of prediction labels | 3.897 | 4.464 |
| Max. # of prediction labels | 14 | 16 |
| Avg. # of links to KG | 8.567 | 8.598 |
| Max. # of links to KG | 22 | 24 |

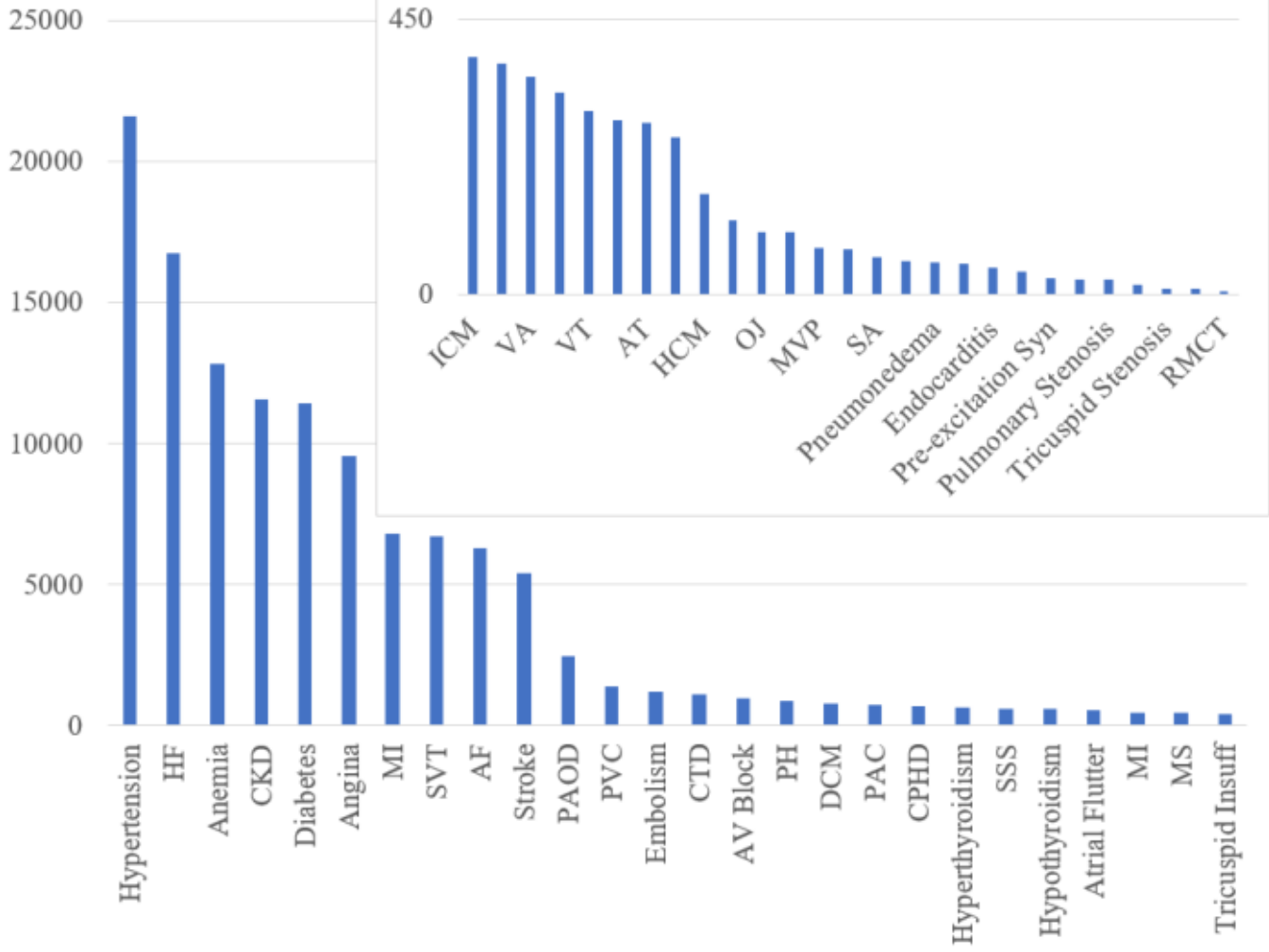

Fig. 3 Label count in the PLAGH dataset

### *B. Data and Preprocessing Procedure*

Two real-world EHR datasets were used to train and evaluate the performance of the proposed model, i.e., the PLAGH dataset and the MIMIC-III dataset [34]. The PLAGH dataset includes EHR data of 84,705 hospitalizations of 27,217 cardiovascular disease patients who visited the Cardiology Department of Chinese PLA General Hospital between 1998 and 2018. The MIMIC-III dataset includes EHR data of 58,997 hospitalizations of 46,519 patients who visited the intensive care unit (ICU) of Beth Israel Deaconess Medical Center between 2001 and 2012.

We implemented basically the same data preprocessing procedure for both datasets. Patients with only one admission were excluded. The EHR data of each hospitalization was reorganized into two non-overlapping parts, i.e., the patient character and patient feature. The patient character is a 65-dimensional binary vector. Each element can be directly mapped into an entity in the KG to indicate whether a patient "connects" the KG at the entity. The patient feature vector records the remaining information in the EHR, i.e., treatments, vital signs, exams, demographics, etc. The features whose missing rate is larger than 30% are excluded. Meanwhile, hospitalizations whose missing rate is larger than 30% or hospitalizations without direct connections to the KG were also

TABLE 2
PREDICTION PERFORMANCE

| Dataset | Model | macro AUC | Top 1 Hit | Top 3 Hit | Top 5 Hit | Top 10 Hit |
|---|---|---|---|---|---|---|
| PLAGH | Proposed Model | 0.739 (0.005) | 0.830 (0.006) | **2.008 (0.017)** | 2.602 (0.018) | 3.052 (0.021) |
| | XGBoost | 0.726 (0.003) | **0.837 (0.005)** | 1.980 (0.008) | 2.630 (0.013) | 3.290 (0.016) |
| | MLP | **0.743 (0.008)** | **0.837 (0.006)** | 1.981 (0.015) | **2.632 (0.015)** | **3.296 (0.019)** |
| | LR | 0.738 (0.006) | 0.832 (0.004) | 1.952 (0.009) | 2.620 (0.011) | 3.292 (0.018) |
| MIMIC | Proposed Model | **0.639 (0.006)** | **0.758 (0.017)** | **1.873 (0.017)** | 2.583 (0.027) | 3.423 (0.031) |
| | XGBoost | 0.619 (0.005) | 0.730 (0.005) | 1.871 (0.022) | **2.730 (0.026)** | 3.663 (0.042) |
| | MLP | 0.637 (0.011) | 0.720 (0.013) | 1.853 (0.046) | 2.711 (0.059) | 3.730 (0.046) |
| | LR | 0.635 (0.010) | 0.716 (0.007) | 1.832 (0.029) | 2.691 (0.041) | **3.732 (0.046)** |

The horizon T of the proposed model is set to two in this experiment.

excluded. All continuous variables in the two datasets were transformed into the standard normal distribution. Finally, we utilized the multiple imputation method to impute all missing data.

Table 1 represents the profile of the two datasets after the data preprocessing procedure. The preprocessed PLAGH dataset contains 30,918 hospitalizations, while the preprocessed MIMIC-III dataset contains 10,693 hospitalizations. We also investigated the label distribution to check whether our dataset is imbalanced. As shown in Figure 3, the top 10 diseases in the PLAGH dataset occupy more than 85% of the labels, demonstrating that our dataset is highly imbalanced. We also observed a similar data imbalance phenomenon in the MIMIC-III dataset, although the detailed label distribution is not reported to save space. This fact indicates the necessity of adding the regularized policy entropy term in our study.

### *C. Experimental Setup*

#### *1) Prediction Task*

We utilized the EHR data of a patient to predict the diseases in his/her next hospitalization. Because we predicted the diseases according to the KG, the prediction task is actually a multilabel prediction task with 53 targets.

#### *2) Metrics*

We mainly used the average area under the receiver operating curve, a.k.a., macro AUC, to evaluate the predictive performance of the proposed model and baselines. Meanwhile, we also compared the top k hit number among models to analyze their practical prediction ability. To fully utilize the EHR data, we adopted the five-fold cross-validation test to evaluate the model and reported the average performance (with standard deviation).

#### *3) Baselines*

Three commonly seen machine learning models in medical research, i.e., multilayer perceptron (MLP), XGBoost [35], and logistic regression (LR), were used as baselines. As XGBoost and LR are not multilabel prediction models, we utilized the one-vs-all framework to allow these two models to accomplish a multilabel prediction task.

#### *4) Experimental Environment*

We performed Bayesian optimization 40 times to find the optimum hyperparameters [36]. All experiments were implemented with Pytorch 1.5.1 and scikit-learn 0.20.2 on a machine equipped with Intel Xeon E5-2640, 16 GB RAM, 2 Nvidia Titan Vs, and CUDA 9.1.

### *D. Prediction Performance*

Table 2 reports the prediction performance of the disease prediction task. Generally, the proposed model achieves similar performance compared to the three baselines. In terms of the macro AUC, the proposed model achieves 0.739 (second place) and 0.639 (first place) in the PLAGH dataset and the MIMIC-III dataset, respectively. Our model obtains the best performance in the top 3 hits for both experimental datasets (PLAGH: 2.008, MIMIC: 1.873), the best performance in the top 1 hit for the MIMIC-III dataset (0.758), and similar performance compared to baselines in the top 1 hit for the PLAGH dataset (0.830). The superiority of the proposed model in the top 1 hit and top 3 hits indicates that it is more likely to assign a larger probability to the disease that truly occurs in the future, demonstrating its potential to be applied in practice. However, the proposed model also has a disadvantage. The top 5 hit number and top 10 hit number of the proposed model are worse than the baselines, which may be attributed to the design and experimental setting. Because we predict the onset of diseases by conducting a path along the KG, we cannot predict de novo diseases that are not correlated to the current diseases or risk factors, and we cannot predict diseases whose progression path is not recorded in the KG. Meanwhile, because we set the horizon of the RL agent to two for the sake of performance, our model will not give any probability to diseases that do not have direct connections to the currently diagnosed disease or risk factors.

### *E. Case Study*

In Figure 4.a, we draw the disease progression paths of a male patient in the PLAGH dataset within a subset of the KG. Of note, we only highlight the transitions with probabilities larger than 0.1 for the sake of clarity. The patient (blue entity) had hypertension, diabetes, anemia, and obesity (red entity). Then, we predict that the patient will develop heart failure, angina, hypertension, and diabetes (red entity) in the future according to the progression paths (red relations) generated by our model. All predicted diseases in the figure are diagnosed in the patient's next admission. Figure 4.b is the simplified version of Figure 4.a, where only the meaningful entities and relations are reserved. The content of the figure can be summarized in three points, and all of them are reasonable from a medical point of view. 1) Hypertension, diabetes, and anemia are the top three predicted diseases, and their probabilities are mainly from the "self-loop" link, indicating that the patient will be diagnosed with the three diseases because he has these diseases at present.

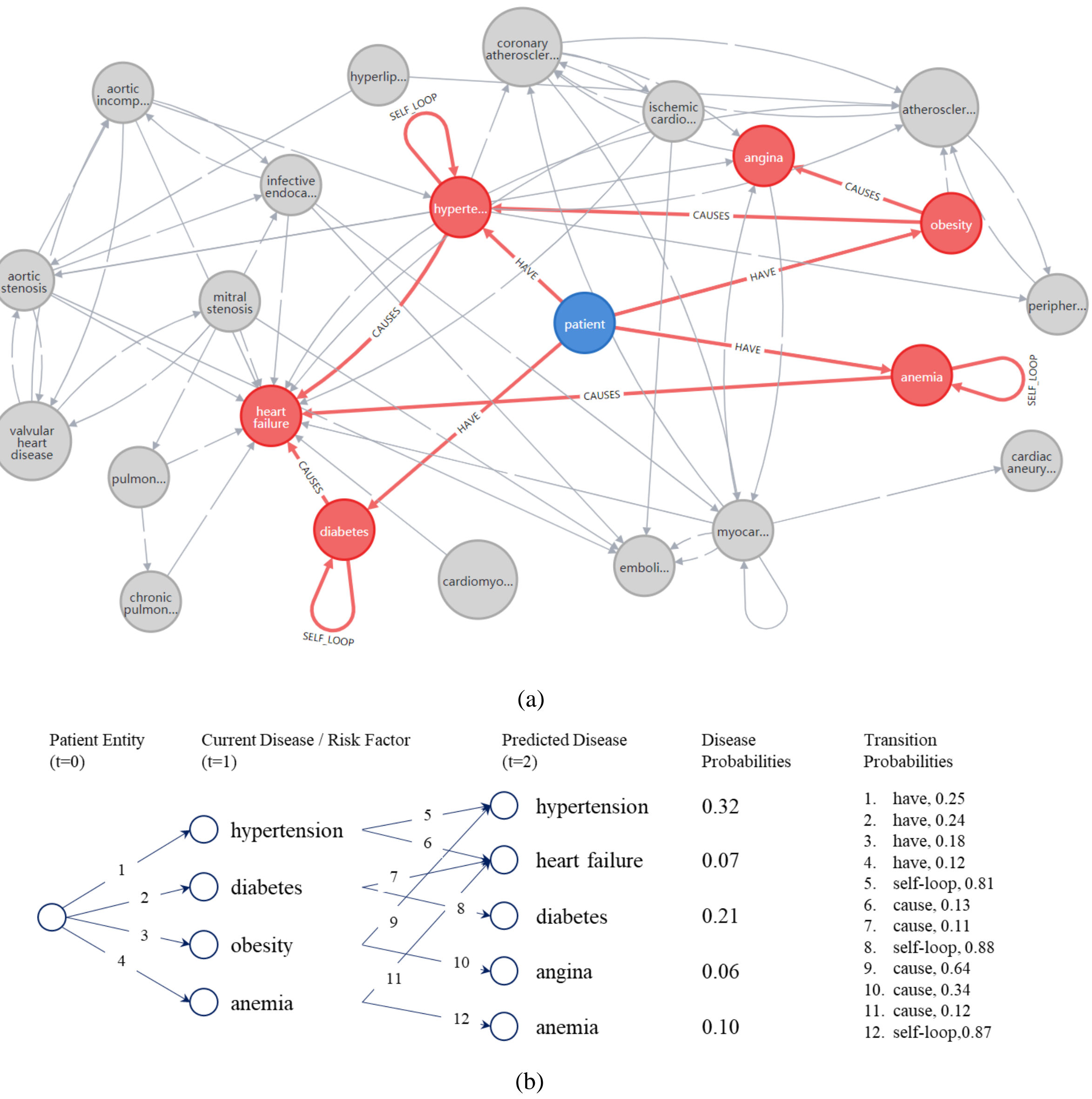

Fig. 4 Example of disease progression paths and probabilities of predicted diseases. (a) The possible disease progression paths of a patient, where the blue entity indicates the patient, the red entity indicates current disease/risk factor or predicted disease, and the red edge presents the disease progression path. For display clarity, we only draw a subset of our KG, and the types of relations that are not in the highlighted disease progression paths are left out (b) We discard unimportant disease entities and relations and show the disease probabilities and transition probabilities.

Because all three diseases are chronic diseases, which are hard to cure, this explanation is reasonable. 2) The given patient is likely to develop heart failure because he has hypertension, anemia, and diabetes. This inference is also reasonable because hypertension and anemia may decrease effective cardiac output, while obesity is an independent risk factor for heart failure [33]. 3) Obesity is the main cause of angina, while diabetes and anemia are not, indicating that our model may have the potential to predict the disease from a pathology perspective and extract the real cause of disease among several possible choices. Of note, we set the horizon of RL to two in consideration of performance. If we set a larger horizon, the model will generate a larger path set consisting of more complex disease progression paths.

### F. *Influence of Horizon T*

The horizon $T$ is a critical parameter in our study. A larger horizon $T$ will encourage the RL agent to explore more diverse and complex disease progression paths at the expense of the risk of going over the target entities. To determine the best choice of the horizon, we evaluate the model with different horizons T (2, 3, 4, or 5), and the result is shown in Figure 5.a. In all metrics, the model with the horizon set to two outperforms the other models significantly in both datasets. Therefore, we set the horizon of the RL agent to two in our experiment.

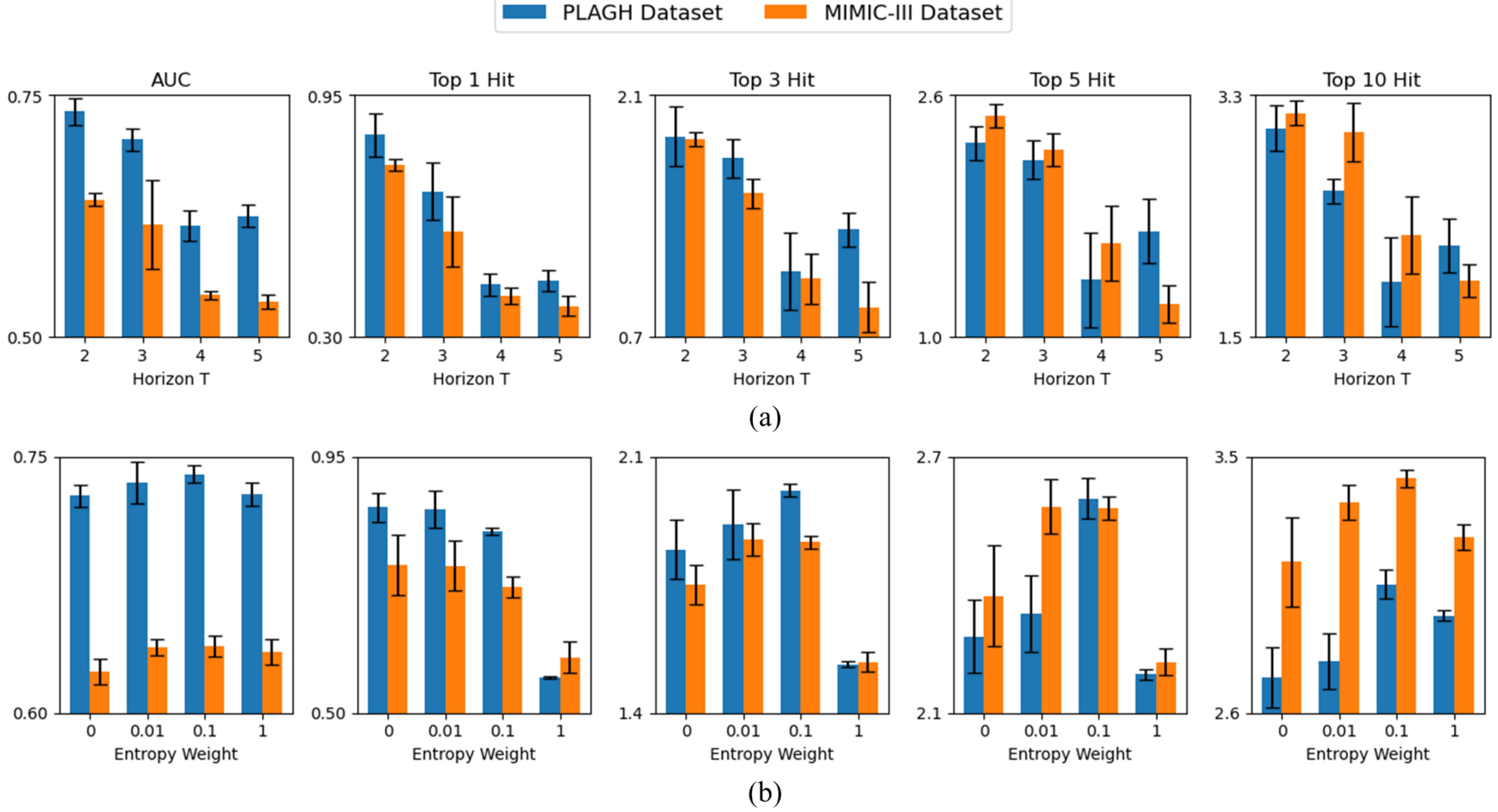

Fig. 5 (a) The performance of models with different horizons T. (b) The performance of models with different weights of the policy entropy. The error bar in the figure is the standard deviation of the results of five-fold cross-validation.

### G. *Influence of Entropy*

We also investigated the influence of the policy entropy term ($\alpha H(\pi)$ in Equation 7) by testing the model's performance with different entropy weights (0, 0.01, 0.1, and 1), whose result is shown in Figure 5.b. The model performance gradually increases with entropy weight in macro AUC, top 3 hits, top 5 hits, and top 10 hits when the entropy weight is less than 0.1. Meanwhile, these metrics will decline rapidly if we keep increasing the entropy weight. The top 1 hit, although it declines monotonically, is relatively insensitive to the change in entropy weight when it is less than 0.1, and it also declines rapidly when the weight is larger than 0.1. The experimental results demonstrate that adding the policy entropy term with an appropriate weight can alleviate the influence of data imbalance and significantly improve model performance. According to the experimental results of the hyperparameter selection process, we finally set the entropy weights to 0.13.

## V. Discussion

The combination of expert knowledge and medical big data has long been regarded as a solution to achieve transparent and reliable disease prediction [5]. However, only a few studies have successfully incorporated medical knowledge into a data-driven model, and none of them can provide intuitive explanations for the prediction result [11]-[14]. Inspired by the idea from the path-based method in recommendation systems, we presented a KG-based learning model to address this challenge. In our model, medical knowledge and medical data play different roles in the prediction task. Medical knowledge, which is reorganized in a KG, is the foundation of the model's interpretability. It provides all possible disease progression paths. Medical data, which are used to train an RL agent, are the foundation of individualized prediction. Once the RL agent is trained, it can generate ubiquitous disease progression paths within the KG according to the medical data of a target patient. To the best of our knowledge, no prior study has endowed a different role to knowledge and data to achieve interpretable prediction. We argue that this design can be a promising direction to leverage the potential of medical knowledge and medical data jointly for specific prediction tasks. The experiments conducted on two real-world datasets demonstrate that our model achieves comparable performance to the state-of-the-art ML-based models. More importantly, the proposed model can generate explainable disease progression paths of patients, which are intuitive for medical diagnosis and thus are advocated by clinicians.

It is worth noting that the entity representation learning process in our study is different from the previous path-based recommendation studies [24], [25]. Previous studies integrated customers, products, and their relations into a large HIN [37], [38], and then adopted a graph embedding learning approach, e.g., TransE [39] and DistMult [40], to learn the distributed representations of entities and relations. Because all customers must be encoded and integrated into the HIN in advance, previous models cannot generate the representations for new customer entities. Meanwhile, they cannot leverage the information that is useful to recommend tasks but cannot be easily mapped to HIN. To alleviate this problem, we separated the representation learning process into independent parts. We adopted the RBM to leverage the cooccurrence characteristics between entities in the representation learning process. The utilization of an autoencoder provides the possibility of fully utilizing the information in EHR. This design, on the one hand, allows us to generate representations for a target patient and dynamically connect the patient entity to the KG at runtime, such that the model can be deployed in real clinical settings. On the other hand, it leverages the information contained in EHR as much as possible and thus can explore the potential of EHR data in a meaningful manner.

As a proof-of-concept study, our model has several limitations that should be noted. 1) We only manually built a small KG that covers essential disease entities and risk factor entities of circulation system diseases. A large number of entities that are related to circulation system disease were not included, e.g., comorbidity entities, pathology entities, and etiology entities. To depict the disease progression path comprehensively and in more detail, we will analyze more medical literature to extend the KG in future studies. 2) In this study, we hypothesize that existing medical studies have fully investigated the relations between disease and risk factor entities. Thereafter, we formulate the disease prediction task as a random walk along with the KG. However, there may be some unknown relations between entities. Our model is not able to predict disease onset that is caused by an unknown disease progression path. Therefore, our model is only appropriate to be applied for diseases that have been extensively explored, e.g., circulation system diseases, and it is not appropriate to be applied in predicting the onset of diseases whose causes are unclear. 3) We utilize the treatment information by including it in the patient feature $\boldsymbol{p_f}$. Although useful, this is a makeshift because we cannot explicitly observe the influences of medical treatments. An ideal model should not only evaluate the positive relations between disease and risk factor entities to generate disease progression paths (like our model) but also evaluate the negative relations between disease and treatment entities to draw the "disease suppression path". To this end, we plan to introduce the negative sampling-based models [41] in our future study to enrich the proposed model in evaluating the positive influence of previous disease and the negative influence of medical treatment simultaneously.

## VI. Conclusion

Leveraging medical knowledge to improve the interpretability of disease prediction models is a fundamental challenge in medical informatics. To address this challenge, we present a new perspective on incorporating medical knowledge into a data-driven model for disease prediction. The proposed model utilizes the RL tactic to conduct interpretable prediction. We carried out extensive experiments on two real-world EHR datasets, and experimental results demonstrated that our model could achieve competitive performance with the baselines for the disease prediction task. More importantly, the interpretable disease progression paths generated by our model were advocated by clinicians.